\documentclass[wcp]{jmlr}

\usepackage{longtable}
\usepackage{booktabs}
\usepackage{longtable}
\usepackage{xcolor}
\usepackage{multirow}
\usepackage{rotating}

\makeatletter
\let\Ginclude@graphics\@org@Ginclude@graphics 
\makeatother

\newtheorem{cor}{Corollary}

\jmlrvolume{}
\jmlryear{2021}
\jmlrworkshop{ACML 2021}

\newif\ifshowcomments
\showcommentsfalse
\ifshowcomments
\newcommand{\mynote}[2]{\fbox{\bfseries\sffamily\scriptsize{#1}}
{\small$\blacktriangleright$\textsf{\emph{#2}}$\blacktriangleleft$}}
\else
\newcommand{\mynote}[2]{}
\fi

\newcommand{\rb}[1]{\textcolor{orange}{\mynote{Robert }{#1}}}
\newcommand{\ak}[1]{\textcolor{green}{\mynote{Aditya }{#1}}}

\newcommand{\system}{DTGAN\xspace}
\newcommand{\systemd}{DTGAN$_\mathcal{D}$\xspace}
\newcommand{\systemg}{DTGAN$_\mathcal{G}$\xspace}
\newcommand{\tgan}{TGAN\xspace}
\newcommand{\tgans}{TGANs\xspace}
\title[\system]{\system: Differential Private Training for Tabular GANs}

  \author{\Name{Aditya Kunar} \Email{A.Kunar@student.tudelft.nl}\\
  \addr TU Delft, Delft, Netherlands
  \AND
  \Name{Robert Birke} \Email{robert.birke@ch.abb.com}\\
  \addr ABB Research Switzerland, Dättwil, Switzerland
  \AND
  \Name{Zilong Zhao} \Email{Z.Zhao-8@tudelft.nl}\\
  \addr TU Delft, Delft, Netherlands
  \AND
  \Name{Lydia Y. Chen} \Email{Y.Chen-10@tudelft.nl}\\
  \addr TU Delft, Delft, Netherlands
 }

\begin{document}

\maketitle
\begin{abstract}
Tabular generative adversarial networks (\tgans) have recently emerged to cater to the need of synthesizing tabular data-- the most widely used data format. While synthetic tabular data offers the advantage of complying with privacy regulations, there still exists a risk of privacy leakage via inference attacks due to interpolating the properties of real data during training. Differential private (DP) training algorithms provide theoretical guarantees for training machine learning models by injecting statistical noise to prevent privacy leaks. However, the challenges of applying DP on \tgan are to determine the most optimal framework (i.e., PATE/DP-SGD) and neural network (i.e., Generator/Discriminator) to inject noise such that the data utility is well maintained under a given privacy guarantee. In this paper, we propose \system, a novel conditional Wasserstein tabular GAN that comes in two variants \systemg and \systemd, for providing a detailed comparison of tabular GANs trained using DP-SGD for the generator vs discriminator, respectively. We elicit the privacy analysis associated with training the generator with complex loss functions (i.e., classification and information losses) needed for high quality tabular data synthesis. Additionally, we rigorously evaluate the theoretical privacy guarantees offered by DP empirically against membership and attribute inference attacks. Our results on 3 datasets show that the DP-SGD framework is superior to PATE and that a DP discriminator is more optimal for training convergence. Thus, we find (i) \systemd is capable of maintaining the highest data utility across 4 ML models by up to 18\% in terms of the average precision score for a strict privacy budget, $\epsilon=1$, as compared to the prior studies and (ii) DP effectively prevents privacy loss against inference attacks by restricting the success probability of membership attacks to be close to 50\%.
\end{abstract}


\section{Introduction}

Tabular GANs have shown promising results in terms of learning the original training data distributions and generating high utility synthetic tabular datasets~\cite{xu2019modeling,zhao2021ctab}. However, utilising privacy sensitive real datasets to train tabular GANs poses a range of privacy issues. Recent studies have shown that GANs may fall prey to membership and attribute inference attacks which greatly endanger the personal information present in the real training data~\cite{gan_leak,priv_mirage}. Therefore, it is imperative to safeguard the training of tabular GANs such that it remains protected against malicious privacy attacks to ensure that synthetic data can be stored and shared across different parties without harm.  

To address the privacy issues in tabular data, prior work~\cite{jordon2018pate,long2019scalable,torkzadehmahani2019dp,torfi2020differentially} relies on Differential Privacy (DP)~\cite{dwork2008differential} for training tabular GANs in a privacy preserving manner. DP is a mathematical framework that provides theoretical guarantees that bounds the statistical difference between any resulting tabular GAN model trained regardless of the existence of any particular individual's information in the original training dataset. Typically, this can be achieved by injecting noise while updating the parameters of a network during back-propagation i.e., DP-SGP~\cite{abadi2016deep,xie2018differentially,chen2020gs}, or by injecting noise while aggregating teacher ensembles using the PATE framework~\cite{papernot2016semi,jordon2018pate}. Moreover, as the tabular GAN is composed of generator and discriminator networks, the prior art has two major pathways for training tabular GANs with differential private guarantees: (i) inject noise while training the discriminator~\cite{xie2018differentially} or (ii) inject noise while training the generator~\cite{chen2020gs}. 

However, existing literature does not provide a well defined consensus on which DP framework (i.e., DP-SGD or PATE) is most optimal for training tabular GANs. And, it further remains unclear which network (i.e., discriminator vs generator) must be trained with DP guarantees to generate samples with higher utility while maintaining strict privacy guarantees. In addition, prior studies rarely investigate the empirical robustness of differential private GANs against privacy attacks. 


In this work, we propose \system, a novel conditional tabular GAN trained with DP-SGD on the Wasserstein loss with gradient penalty~\cite{gulrajani2017improved}. \system has two variants where \systemd and \systemg either train the discriminator or generator with DP-SGD, respectively. Moreover, for \systemg, we precisely account for the privacy cost associated with 2 additional loss functions, namely information loss and classification loss, which are used to improve training of the generator network~\cite{park2018data}. Additionally, to further reduce the privacy cost, we adopt the subsampling technique~\cite{wang2019subsampled} to train the discriminator at \systemd and the generator at \systemg, respectively.  

Through our extensive experimental analysis using state-of-the-art DP tabular GAN baselines on 3 datasets and 4 ML models, we show that DP-SGD outperforms the PATE framework.  Moreover, we highlight the theoretical and practical considerations for training the discriminator vs generator with DP-SGD. We show that \systemd produces the best overall data utility by improving upon prior work by 18\% in terms of average precision score under a strict privacy budget, $\epsilon=1$. Furthermore, a rigorous empirical evaluation against privacy attacks reveals that DP guarantees reduce the risk of privacy loss against inference attacks, by effectively limiting the membership attack success rates to approximately 50\%.

Our contributions can be summarized as:
\begin{itemize}
    \item We propose \system a conditional tabular Wasserstein GAN which marries superior quality of synthesized data stemming from complex loss functions with proven privacy guarantees stemming from DP.
    \item We develop \systemd and \systemg, two variants of \system, to study the optimal placement of privacy enforcement.
    \item We extensively evaluate the two variants against 3 baselines on multiple ML utility metrics assessing the performance of models trained on real and generated data, as well as privacy resilience metrics assessing the difficulty to perform membership and attribute inference attacks.
     
\end{itemize}

The rest of this paper is organized as follows: Sec.~\ref{sec:background} provides a brief background on differential privacy followed by an overview of related work. The two main approaches used to employ differential privacy in T-GAN are elucidated in Sec.~\ref{sec:system}. Sec.~\ref{sec:experiments}, provides a rigorous empirical examination of \system. Finally, Sec.~\ref{sec:conclusion} concludes the paper. 

\section{Background and Related Work}
\label{sec:background}
This section presents the notions pertaining to generative adversarial networks, differential privacy and differential private DNN training techniques relevant to this work.

\subsection{Generative Adversarial Networks}
\label{ssec:gan}

GAN is a popular algorithm to train generative models. 
GANs leverage two opposing models: a generator $\mathcal{G}$ and a discriminator $\mathcal{D}$. The discriminator aims at distinguishing real data from fake data synthesized by the generator. The generator aims at synthesizing data which is indistinguishable from real data to fool the discriminator. The two models are trained together via an adversarial min-max game minimizing the loss of the generator while maximizing the loss of the discriminator.

\subsection{Differential Privacy}
\label{ssec:dp}
DP is the golden standard in privacy. DP protects against privacy attacks by  minimizing the influence of any individual data under a privacy budget. We leverage the R\'enyi Differential Privacy (RDP)~\cite{mironov2017renyi} as it provides stricter bounds on the privacy budget. A randomized mechanism \( \mathcal{M} \) is $(\lambda,\epsilon)$-RDP with order $\lambda$, if 
$D_{\lambda}(\mathcal{M}(S)||\mathcal{M}(S')) = \frac{1}{\lambda-1}log\mathbb{E}_{x\sim\mathcal{M}(S)} \left[  \left(\frac{P[\mathcal{M}(S)=x]}{P[\mathcal{M}(S')=x]} \right) \right]^{\lambda-1}\leq\epsilon$
holds for any adjacent datasets $S$ and $S'$, where
$D_\lambda(P||Q)=\frac{1}{\lambda-1}log\mathbb{E}_{x\sim Q}[(P(x)/Q(x))^\lambda]$ represents the R\'enyi divergence. In addition, a $(\lambda,\epsilon)$-RDP mechanism \( \mathcal{M} \) can be expressed as:
\begin{equation}
    \label{eq:rdp-to-dp}
    (\epsilon+\frac{log1/\delta}{\lambda-1},\delta)\text{-DP}.
\end{equation}
For the purposes of this work $\mathcal{M}$ corresponds to a tabular GAN model and $(\lambda, \epsilon)$ represents the privacy budget.
RDP is a strictly stronger privacy definition than DP as it provides tighter bounds for tracking the cumulative privacy loss over a sequence of mechanisms via the Composition theorem~\cite{mironov2017renyi}. Let $\circ$ denote the composition operator. For \( \mathcal{M}_{1} \),...,\( \mathcal{M}_{k} \) all $(\lambda,\epsilon_i)$-RDP, the composition \( \mathcal{M}_{1} \)$\circ ... \circ$\( \mathcal{M}_{k} \) is
\begin{equation}
    \label{eq:composition}
    (\lambda,\sum_{i}\epsilon_{i})\text{-RDP}.
\end{equation}

\cite{dwork2014algorithmic} defines a Gaussian Mechanism \( \mathcal{M}_{\sigma} \) parameterized by $\sigma$ as:
\begin{equation}
    \label{eq:gaussian_mechanism}
    \mathcal{M}_{\sigma}(x) = f(x) + \mathcal{N}(0,\sigma^{2}I)
\end{equation}
where $f$ denotes an arbitrary function with sensitivity $\Delta_{2}f = \max_{S,S'}||f(S) - f(S')||_{2}$ over all adjacent datasets $S$ and $S'$, and  $\mathcal{N}$ a Gaussian noise with mean 0 and covariance $\sigma^{2}I$.
$\mathcal{M}_{\sigma}$ is $(\lambda,\frac{\lambda\Delta_{2}f^{2}}{2\sigma^{2}})$-RDP~\cite{mironov2017renyi}.

Two more theorems are important for this work.
Post Processing~\cite{dwork2014algorithmic} states that if \( \mathcal{M} \) satisfies $(\epsilon,\delta)$-DP, $F\circ\mathcal{M}$ will satisfy $(\epsilon,\delta)\text{-DP}$.
Hence, it suffices to train one of the two networks in the GAN architecture with DP guarantees to ensure that the overall GAN is compatible with differential privacy.
RDP for Subsampled Mechanisms~\cite{wang2019subsampled} computes the reduction in privacy cost when sub-sampling private data. Formally, let $\mathcal{X}$ be a dataset with $n$ data points and \textbf{subsample} return $m \leq n$ subsamples without replacement from $\mathcal{D}$ (subsampling rate $\gamma = m/n$).
For all integers $\lambda \geq 2 $, if a randomized mechanism $\mathcal{M}$ is $(\lambda,\epsilon(\lambda))$-RDP, then $\mathcal{M}\circ\textbf{subsample}$ is
\begin{equation}
    \label{eq:subsampling}
    (\lambda,\epsilon'(\lambda))\text{-RDP}
\end{equation}

where
$\epsilon'(\lambda) \leq \frac{1}{\lambda-1}log(1 + \gamma^{2}\binom{\lambda}{2}\min\left\{4(e^{\epsilon(2)}-1),e^{\epsilon(2)}\min\{2,(e^{\epsilon(\infty)}-1)^{2}\}\right\} \allowbreak$ + \\$\sum_{j=3}^{\lambda}\gamma^{j} \binom{\lambda}{j}e^{(j-1)\epsilon(j)}\min\{2,(e^{\epsilon(\infty)-1)^{j})}\})$

\subsection{Training with Differential Privacy Guarantees}
\label{ssec:dp-sgd}

DP-SGD~\cite{abadi2016deep} forms the central framework to provide differential privacy guarantees in this work.
DP-SGD uses noisy stochastic gradient descent to limit the influence of individual training samples $x_i$. After computing the gradient $g(x_i)$, the gradient is clipped based on a clipping parameter $C$ and its L2 norm  $\bar{g}(x_i) \leftarrow g(x_i)/\max(1,\frac{||g(x_i)||_{2}}{C})$ and Gaussian noise is added $\Tilde{g}(x_i)\leftarrow \bar{g}(x_i)+\mathcal{N}(0,\sigma^{2}C^{2}I))$. $\Tilde{g}$ is then used in place of $g$ to update the network parameters as in traditional SGD. $\Tilde{g}$ fits the definition of a Gaussian mechanism $\mathcal{M}_{\sigma}$.

One of the biggest challenges with DP-SGD is tuning the clipping parameter $C$ 
since clipping greatly degrades the information stored in the original gradients~\cite{chen2020gs}. Choosing an optimal clipping value that does not significantly impact utility is crucial. However, tuning the clipping parameter is laborious as the optimal value fluctuates depending on network hyperparameters (i.e. model architecture, learning rate)~\cite{abadi2016deep}.
To avoid an intensive hyper-parameter search, \cite{chen2020gs}~proposes to use the Wasserstein loss with a gradient penalty term.  
This term ensures that the discriminator generates bounded gradient norms which are close to 1 under real and generated distributions. Therefore, an optimal clipping threshold of $C=1$ is obtained analytically.

\subsection{Differential Private Tabular GANs}


To avoid leaking sensitive information on single individuals, related work explored several possibilities to provide differential privacy guarantees with GANs. Table~\ref{table:DD} provides an overview across the main ingredients used. Both of the two private learning techniques have been applied. \cite{jordon2018pate} uses PATE~\cite{papernot2016semi} which relies on output sanitization perturbing the output of an ensemble of teacher discriminators via Laplacian noise to train a student discriminator scoring the generated samples from the generator. One key limitation is that the student discriminator only sees synthetic data. Since this data is potentially unrealistic the provided feedback can be unreliable. \cite{xie2018differentially,chen2020gs,torfi2020differentially} use DP-SGD coupled with the Wasserstein loss. Moreover, \cite{xie2018differentially} uses a momentum accountant whereas \cite{chen2020gs,torfi2020differentially} a R\'enyi differential privacy accountant. 
The Wassertein loss is known to be more effective against mode-collapse as compared to the KL divergence~\cite{pmlr-v70-arjovsky17a}. The RDP accountant provides tighter bounds on the privacy costs improving the privacy-utility tradeoff.
To incorporate differential privacy guarantees and make the training compatible with the Wasserstaien Loss~\cite{xie2018differentially,torfi2020differentially} use weight clipping to enforce the Lipschitz constraint. The drawback is the need for careful tuning of the clipping parameter, see Sec.~\ref{ssec:dp-sgd}. To overcome this issue, \cite{chen2020gs} enforces the Lipschitz constraint via a gradient penalty term as suggested by~\cite{gulrajani2017improved}, but addresses only images which are a better fit for GANs and studies it's efficacy only for training the generator network. 



The proposed \system leverages RDP-based privacy accounting as well as the Wasserstein loss with a gradient penalty for improved privacy accounting and training stability, respectively. However, different from the related work, we tackle differential privacy at both the generator, \systemg, and discriminator, \systemd, to investigate the most optimal site for integrating DP in tabular GANs. Moreover, we elucidate the privacy cost associated with complex loss functions (i.e., classification and information losses) used by the generator for enhancing the quality of synthetic tabular datasets. Additionally we experimentally validate not only the GAN performance but also the resilience against membership and inference attacks.

\begin{table}[t]
\caption{Overview of related work.}
\resizebox{1.0\columnwidth}{!}{
\centering
\begin{tabular}{|c|c|c|c|c|c|c|c|c|}
\hline
\multicolumn{2}{|c|}{\textbf{Model}} & \textbf{DP Algo} & \textbf{Loss$^*$} &\textbf{DP Site} & \textbf{Noise}& \textbf{Accountant} & \textbf{Data Format}\\ 
\hline
\multicolumn{2}{|c|}{PATE-GAN~\cite{jordon2018pate}} & PATE & KL Divergence  &  $\mathcal{D}$   & Laplacian  & PATE & Table\\\hline
\multicolumn{2}{|c|}{DP-WGAN~\cite{xie2018differentially}} & DP-SGD & Wasserstein + WC    &  $\mathcal{D}$   & Gaussian  & Moment & Image $\&$ Table \\ \hline
\multicolumn{2}{|c|}{GS-WGAN~\cite{chen2020gs}} & DP-SGD & Wasserstein + GP & $\mathcal{G}$  & Gaussian & RDP & Image\\ \hline

\multicolumn{2}{|c|}{RDP-GAN~\cite{torfi2020differentially}} & DP-SGD & Wasserstein + WC & $\mathcal{D}$  & Gaussian & RDP & Table\\ \hline

\multirow{2}{*}[-0.1em]{\begin{sideways}\bf Ours\end{sideways}} 
& \systemd & DP-SGD & Wasserstein + GP   &  $\mathcal{D}$  & Gaussian  & RDP  & Table\\
\cline{2-8} 
& \systemg & DP-SGD & Wasserstein + GP    &  $\mathcal{G}$   & Gaussian  & RDP  & Table \\ \hline
\end{tabular}
}
\footnotesize $^*$WC (Weight Clipping), GP (Gradient Penalty)
\label{table:DD}
\end{table}

\section{\system}
\label{sec:system}

\system is a novel approach to generate tabular datasets with strong DP guarantees. It utilizes the DP-SGD framework~\cite{abadi2016deep}
and the subsampled RDP moments accountant technique~\cite{mironov2017renyi,wang2019subsampled}
to preserve privacy and account for the cost, respectively. In addition, it makes use of the Wasserstein loss with gradient penalty 
~\cite{gulrajani2017improved}
to effectively bound the gradient norms thereby providing an analytically derived optimal clipping value for better preserving gradient information after being clipped in regards to DP-SGD as shown in the work of~\cite{chen2020gs}.

Thus, in Sec.~\ref{subsec:training_objectives}, we first illustrate the training objectives used by the discriminator and generator respectively. Next, we present the implementation and privacy analysis of two variants of \system. Sec.~\ref{ssec:dtgand} details training of the discriminator network with DP guarantees whereas Sec.~\ref{ssec:dtgang} describes applying private learning on the generator network. Both approaches are studied to obtain the most optimal configuration for training \system.

\subsection{Training Objectives}
\label{subsec:training_objectives}

We change the interpolation method in the Wasserstein with gradient penalty loss function. Originally the authors define the input, i.e. random samples $\hat{x}\sim\mathbb{P}_{\hat{x}}$, as sampling along the straight lines between pair of points sampled from the original data distribution $\mathbb{P}_{r}$ and the generator distribution $\mathbb{P}_{g}$. However, this relies on the assumption that data points form a uniformly distributed hypercube. Since this assumption may not always hold in practice, spherical interpolates~\cite{shoemake1985animating} are used in this work for accounting the possible curvature of the latent space. This was found to yield better data utility in preliminary experiments. Therefore, the training objectives $\mathcal{L}_D$ and $\mathcal{L}_G$ for the discriminator and generator are expressed as:
\begin{equation}
    \mathcal{L}_{D} = \underbrace{\mathbb{E}_{\Tilde{x}\sim\mathbb{P}_g}[D(\Tilde{x})]-\mathbb{E}_{x\sim\mathbb{P}_{r}}[D(x)]}_{\text{Wasserstein loss}}+\underbrace{\tau\mathbb{E}_{\hat{x}\sim\mathbb{P}_{\hat{x}}}[(||\triangledown_{\hat{x}}D(\hat{x})||_{2}-1)^{2}]}_{\text{Gradient penalty}}
\end{equation}
\begin{equation}
    \label{eq:generatorloss}
    \mathcal{L}_{G}= -\mathbb{E}_{\Tilde{x}\sim\mathbb{P}_g}[D(G(\Tilde{x}))]
\end{equation}
where $D$ is the set of 1-Lipschitz functions defining the discriminator network, $G$ represents the generator network and $\tau$ is the penalty coefficient.

The generator of \system utilizes two additional loss terms added to \eqref{eq:generatorloss}, known as the classification and information loss, to generate tabular data with greater fidelity than the original~\cite{park2018data}. The classification loss $\mathcal{L}_{C} = \mathbb{E}[|l(\Tilde{x})-C(fe(\Tilde{x}))|]_{\Tilde{x}\sim\mathbb{P}_g}$ corresponds to training $G$ using the added classifier network, i.e. $C$, where $l(.)$ is a function that returns the class label of any given data row and $fe(.)$ deletes the class feature of that data row. It quantifies the discrepancy between synthesized and predicted class labels as outputted by an additional classifier module trained to generate predictions using real data. This helps to increase the semantic integrity of synthetic records.
The information loss 
$\mathcal{L}_{I}= \mathcal{L}_{mean} + \mathcal{L}_{sd}$ where $\mathcal{L}_{mean} = ||\mathbb{E}[f_x]_{x\sim\mathbb{P}_{r}} - \mathbb{E}[f_{\Tilde{x}}]_{\Tilde{x}\sim\mathbb{P}_g}||_{2}$ and $\mathcal{L}_{sd} = ||\mathbb{SD}[f_x]_{x\sim\mathbb{P}_{r}} - \mathbb{SD}[f_{\Tilde{x}}]_{\Tilde{x}\sim\mathbb{P}_g}||_{2}$ penalizes the discrepancy between statistics of the generated data and the real data and prevents the generator to over-train on the current discriminator~\cite{salimans2016improved}. $f_x$ and $f_{\Tilde{x}}$ denote the resulting features obtained from the penultimate layer of $D$ for a real and generated sample and $\mathbb{E}$ and $\mathbb{SD}$ denote the mean and standard deviations of the features, respectively. Lastly, the generator loss is a by-product of the conditional architecture and is simply used to ensure that the generator respects the conditional constraints imposed during training~\cite{xu2019modeling}. 

\subsection{DP-Discriminator}
\label{ssec:dtgand}

The first variant, \systemd, trains the discriminator using differential private-SGD where the number of training iterations is determined based on the total privacy budget $(\epsilon$,$\delta)$. Thus, to compute the number of iterations, the privacy budget spent for every iteration must be bounded and accumulated. 
For this purpose we use the subsampled RDP analytical moments accountant technique. 
The theoretical analysis of the privacy cost is presented below:

\begin{cor}
Each discriminator update satisfies $(\lambda,2B\lambda/\sigma^{2})$-RDP where $B$ is the batch size.
\end{cor}

\begin{proof}
Let $f=clip({\bar{g}_D},C)$ be the clipped gradient of the discriminator before adding noise. The sensitivity is derived via the triangle inequality:
\begin{equation}
    \Delta_{2}f = \max_{S,S'}||f(S)-f(S')||_{2} \leq 2C
\end{equation}

Since $C=1$ as a consequence of the Wasserstein loss with gradient penalty 
and by using \eqref{eq:gaussian_mechanism} the DP-SGD procedure denoted as $\mathcal{M}_{\sigma,C}$ parameterized by noise scale $\sigma$ and clipping parameter $C$ may be represented as being $(\lambda,2\lambda/\sigma^{2})$-RDP. 

Furthermore, each discriminator update for a batch of real data points $\{x_i,..,x_B\}$ can be represented as 

\begin{equation}
    \Tilde{g}_D = \frac{1}{B}\sum_{i=1}^{B}\mathcal{M}_{\sigma,C}(\triangledown_{\theta_D}\mathcal{L}_{D}(\theta_D,x_i))
\end{equation}

where $\tilde{g}_{D}$ and $\theta_{D}$ represent the perturbed gradients and the weights of the discriminator network, respectively. This may be regarded as a composition of $B$ Gaussian mechanisms and treated via \eqref{eq:composition}. The privacy cost for a single gradient update step for the discriminator can be expressed as $(\lambda,\sum_{i=1}^{B}2\lambda/\sigma^{2})$ or equivalently $(\lambda,2B\lambda/\sigma^{2})$.
\end{proof}

Note that $\mathcal{M}_{\sigma,C}$ is only applied for those gradients that are computed with respect to the real training dataset~\cite{abadi2016deep,zhang2018differentially}. Hence, the gradients computed with respect to the synthetic data and the gradient penalty term are left undisturbed. Next, to further amplify the privacy protection of the discriminator, we rely on~\eqref{eq:subsampling} with subsampling rate $\gamma=B/N$ where $B$ is the batch size and $N$ is the size of the training dataset. Intuitively, subsampling adds another layer of randomness and enhances privacy by decreasing the chances of leaking information about particular individuals who are not included in any given subsample of the dataset. 
Lastly, it is worth mentioning that the Wasserstein loss with gradient penalty training objective has one major pitfall with respect to the privacy cost. This is because, it encourages the use of a stronger discriminator network to provide more meaningful gradient updates to the generator. This requires performing multiple updates to the discriminator for each corresponding update to the generator leading to a faster consumption of the overall privacy budget. 

\subsection{DP-Generator}
\label{ssec:dtgang}
 
\begin{figure}[t]
  \centering
  \begin{minipage}[b]{0.46\textwidth}
	\begin{center}
        \includegraphics[width=1\columnwidth]{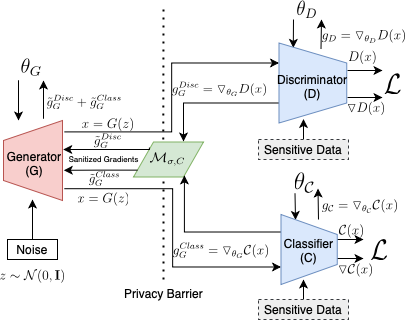}
		\caption{Privacy preserving generator}
		\label{fig:Gen}
	\end{center}
  \end{minipage}
  \hfill
  \begin{minipage}[b]{0.46\textwidth}
 \includegraphics[width=1\columnwidth]{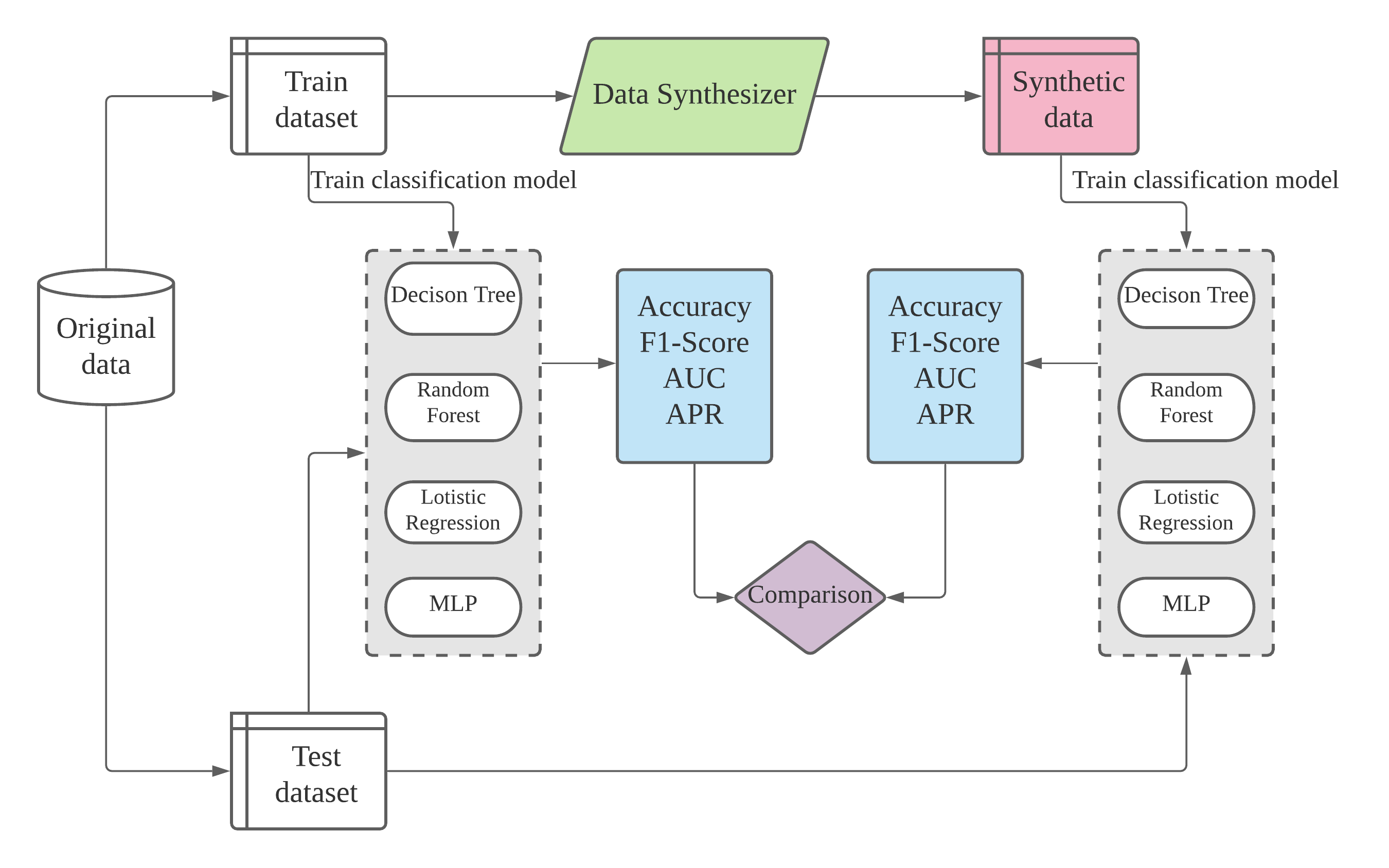}
		\caption{\centering Evaluation flows for ML utility}
    \label{fig:settingA}
  \end{minipage}
\end{figure}

The second variant, \systemg, trains the generator network with DP guarantees. Fig.~\ref{fig:Gen} depicts the process. The gradients from the discriminator and classifier (i.e., $\tilde{g}_{G}^{Disc}$ and $\tilde{g}_{G}^{Class}$) are selectively perturbed via the same $\mathcal{M}_{\sigma,C}$ DP-SGD procedure which is used for updating the generator's weights. 
The selective perturbation of the gradients is necessary as the combined training objective of the generator (i.e., information, classifier and generator loss) does not entirely depend on the original training data. As an example, consider the generator loss which is only used to ensure that the generated data exactly matches the constraint given by the conditional vector sampled randomly during training and as a result, is independent of the real training data~\cite{xu2019modeling}. 
With this in mind we present the privacy analysis for training the generator via DP-SGD and the subsampled RDP moments accounting. 

\begin{cor}
Each generator update satisfies $(\lambda,6B\lambda/\sigma^{2})$-RDP where $B$ is the batch size.
\end{cor}
\begin{proof}
Let $f_{Disc}=clip({\bar{g}_G^{Disc}},C)$ be the clipped gradient of the generator computed with respect to $\mathcal{L}_{G}$ before adding noise. The sensitivity is derived via the triangle inequality:
\begin{equation}
    \Delta_{2}f_{Disc} = \max_{S,S'}||f_{Disc}(S)-f_{Disc}(S')||_{2} \leq 2C
\end{equation}

Since as before $C=1$ and by using~\eqref{eq:gaussian_mechanism} the randomized mechanism $\mathcal{M}_{\sigma,C}$ may similarly be represented as being $(\lambda,2\lambda/\sigma^{2})$-RDP. 

However, due to the addition of the information loss denoted as $\mathcal{L}_{I}$, the generator requires an additional fetch of gradients from the discriminator (i.e., $g_G^{Disc}$) computed with respect to $\mathcal{L}_{I}$ which in turn doubles the number of times $\mathcal{M}_{\sigma,C}$ is applied. Note that the sensitivity remains the same leading to an identical privacy cost (i.e., $(\lambda,2\lambda/\sigma^{2})$-RDP).

Likewise for the classifier loss expressed as $\mathcal{L}_{C}$, let $f_{Class}=clip({\bar{g}_G^{Class}},C)$ be the clipped gradient of the generator back-propagated from the classifier before adding noise. The sensitivity is similarly derived via the triangle inequality:

\begin{equation}
    \Delta_{2}f_{Class} = \max_{S,S'}||f_{Class}(S)-f_{Class}(S')||_{2} \leq 2C
\end{equation}

For ease of derivation, the clipping parameter for the classifier module is also, $C=1$. By using~\eqref{eq:gaussian_mechanism} once again, $\mathcal{M}_{\sigma,C}$ is $(\lambda,2\lambda/\sigma^{2})$-RDP. 

Thus, to do a single update of the generator's weights $\theta_{G}$, the randomized mechanism $\mathcal{M}_{\sigma,C}$ is first applied twice for the discriminator network and once more for the classifier network with a fixed privacy cost of $(\lambda,2\lambda/\sigma^{2})$-RDP. Formally, this can be expressed as

\begin{equation}
    \tilde{g}_G = \sum_{i=1}^{L} \mathcal{M}_{\sigma,C}(\triangledown_{\theta_G}\mathcal{L}_{i}(\theta_G))
\end{equation}

where $L$ represents the set of losses for which the gradients are computed (i.e., \{${\mathcal{L}_{G}, \mathcal{L}_{I}, \mathcal{L}_{C}}$\}) and $\tilde{g}_{G}$ $\&$ $\theta_{G}$ represents the perturbed gradients and the weights of the generator network, respectively. This sequence can once again be interpreted as a composition of Gaussian mechanisms which allows the use of \eqref{eq:composition} to express the cost for an individual data point as $(\lambda,\sum_{i=1}^{3}2\lambda/\sigma^{2})$-RDP. And, the privacy cost for a batch of data points $\{x_i,..,x_B\}$ can be similarly extended to be   $(\lambda,\sum_{i=1}^{B}\sum_{i=1}^{3}2\lambda/\sigma^{2})$ or equivalently $(\lambda,6B\lambda/\sigma^{2})$. 
\end{proof}

Next, to amplify the privacy protection for the generator, \eqref{eq:subsampling}~is analogously used. However, in this case, the original training dataset is divided into disjoint subsets of equal size where a unique discriminator is trained for each subset independently. The size of each subsampled data is defined as $N_{d}/N$ where $N_{d}$ is the total number of discriminators and $N$ is the size of the full training dataset. Thus, during training, one out of the total number of discriminators is chosen randomly for every iteration to provide gradient updates to the generator on the basis of the corresponding subsampled dataset. In this way, the subsampling rate for the generator is defined to be $\gamma = 1/N_{d}$ .

Unfortunately, training multiple discriminators on smaller subsamples is problematic due to the lack of enough training iterations for any given discriminator in comparison to the generator. Moreover, reducing the number of samples via subsampling increases the potential of overfitting a discriminator on its respective subsample. \cite{chen2020gs}~recommends to alleviate the first problem by pre-training multiple discriminator networks with a standard generator without DP. since this does not breach the DP guarantees for the generator. However, in practice, the results were not found to be affected by the presence of pre-trained discriminators in preliminary experiments.

Lastly, \eqref{eq:rdp-to-dp} is used to convert the overall cumulative privacy cost computed in terms of RDP back to $(\epsilon,\delta)$-DP for both approaches.

\section{Experimental Analysis}
\label{sec:experiments}

\subsection{Experimental Setup}

\begin{table}[t]
\centering
\caption[DD]{Description of Datasets.}
\resizebox{0.95\columnwidth}{!}{
\begin{tabular}{ |c|c|c|c|c|c|c|c| }
\hline
\textbf{Dataset} & \textbf{Train/Test Split} &\textbf{Target Variable} & \textbf{$\mbox{Continuous}$}  & \textbf{$\mbox{Binary}$} & \textbf{$\mbox{Multi-class}$} \\ 
\hline
Adult     & 39k/9k   & ``income''        & 3   & 2  & 7 \\
\hline
Credit    & 40k/10k  & ``Class''         & 30  & 1 & 0  \\
\hline
Loan      & 4k/1k    & ``PersonalLoan''  & 5   & 5  & 2  \\
\hline
\end{tabular}
}
\label{table:DDE}
\end{table}


{\bf Datasets.} We evaluate \system using three well-known datasets, i.e., Adult (\cite{UCIdataset}), Credit (\cite{kagglecredit}) and Loan (\cite{kaggleloan}). Tab.~\ref{table:DDE} highlights the salient features of each dataset.
\\
{\bf Baselines}. Both variants of \system are compared against 2 state-of-the-art architectures: PATE-GAN~\cite{jordon2018pate} and DP-WGAN~\cite{xie2018differentially}.
For PATE-GAN we use the original code and parameters provided by the authors. Similarly for DP-WGAN, except that we change the structure of the neural networks to match the ones from PATE-GAN to compare DP-SGD vs PATE frameworks evenly. Note that to compute privacy cost fairly, the RDP accountant is used for all approaches that use DP-SGD as it provides tighter privacy guarantees than the moment accountant~\cite{wang2019subsampled}. Lastly, we also feature \system with no privacy budget (i.e., $\epsilon=\infty$) denoted as \tgan. 
\\
{\bf \system}.
\system is trained with a gradient penalty coefficient, $\tau=10$ as recommended by~\cite{gulrajani2017improved}. Batch size is 64 for all models. All other parameters use their default value. 
\ak{I don't know what else specifics to add, the rest of the hyper-parameters like learning rate etc is not so relevant and I do not tune hyper-parameters of any of the algorithms and use default values where-ever necessary. Batch size is 64 for all baselines. The appendix highlights some additional hyper-parameters for PATE-GAN and DP-WGAN. } \rb{from were do you get the defualt values?} \rb{it is not so much important that you tune them or not, it is rather to know the values to be able to repeat the experiments.}

\subsection{Evaluation Metrics}
We evaluate \system from multiple points of view. On the one hand we check the GAN performance in terms of how similar and useful the generated samples are. On the other hand we evaluate the resilience against two types of privacy attacks.

\subsubsection{GAN Performance}

{\bf Machine Learning Utility}.
\label{Ch3:ml_efficacy}
To quantify the ML utility, we compare the performance achieved by 4 widely used machine learning algorithms on the real versus the synthetic data: decision tree classifier, random forest classifier, multinomial logistic regression and multi-layer perceptron (MLP). We use Python and scikit-learn 0.24.2. 
All model hyper-parameters use their default value. Lastly, we apply Min-Max normalisation as pre-processing step.

First we split the original data into training and test sets (see Fig.~\ref{fig:settingA}). The training set is used as real data to train the GAN models. Once the training is finished, we use it to synthesize data with the same size as the training set. The synthetic and real training datasets are then used to train pairs of model instances of the 4 machine learning models listed above. We use the test set to compute four performance scores: accuracy, F1-score, area under the ROC (AUC), and average precision score (APR).
The ML utility is measured via the difference in each performance score between each model pair trained on the real and synthetic data.
\\
{\bf Statistical Similarity}. Three metrics are used to quantitatively measure the statistical similarity between the real and synthetic data.
\\
\textit{Jensen-Shannon divergence (JSD)}~\cite{jsd}. JSD provides a measure to quantify the difference between the probability mass distributions of individual categorical variables belonging to the real and synthetic datasets, respectively. Moreover, this metric is symmetric and bounded between 0 and 1 allowing for an easy interpretation of results.
\\
\textit{Wasserstein distance (WD)}~\cite{wgan_test}. In similar vein, the Wasserstein distance is used to capture how well the distributions of individual continuous/mixed variables are emulated by synthetically produced datasets in correspondence to real datasets. We use WD because we found that the JSD metric is numerically unstable for evaluating the quality of continuous variables, especially when there is no overlap between the synthetic and original datasets. 
Note that the WD is calculated after performing a min-max normalisation. 
This helps in making the wasserstein distances comparable across columns with drastically varying scales.
\\
\textit{Difference in pair-wise correlation (Diff. Corr.)}. 
To evaluate how well feature interactions are preserved in the synthetic datasets, we compare the pair-wise correlations matrices. 
We use Pearson correlation coefficient for continuous columns, Theil uncertainty coefficient for categorical columns and correlation ratio between categorical and contnuous variables using the dython library\footnote{\url{http://shakedzy.xyz/dython/modules/nominal/\#compute\_associations}}. Finally, the l2 norm of the difference between the pair-wise correlation matrices for the real and synthetic datasets is computed. 


\subsubsection{Privacy Attacks} 

\textbf{Membership Inference Attack}~\cite{chen2020gan}. It is a binary classification problem in which an attacker tries to predict if a particular target data point $t$ has been used to train a victim generative model. This work assumes that the attacker only needs access to a black-box tabular GAN model, a reference dataset $\mathcal{R}$ and $t$ for which the inference must be made~\cite{priv_mirage}. 

\begin{figure}[t]
  \centering
  \begin{minipage}[b]{0.52\textwidth}
	\begin{center}
    \includegraphics[width=1\columnwidth]{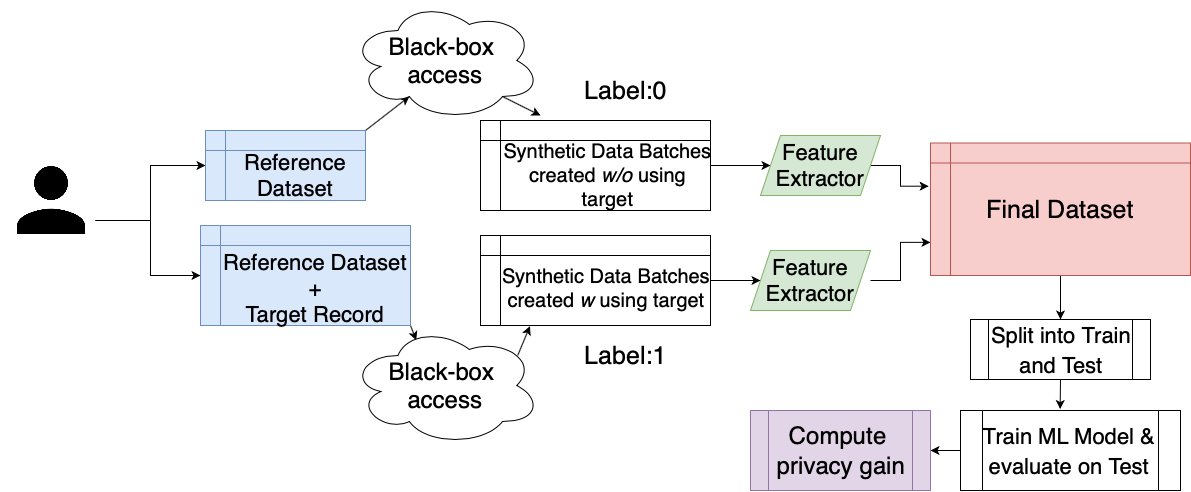}
    \caption{Membership inference attack}
    \label{fig:MIE}
	\end{center}
  \end{minipage}
  \hfill
  \begin{minipage}[b]{0.44\textwidth}
    \includegraphics[width=1\columnwidth]{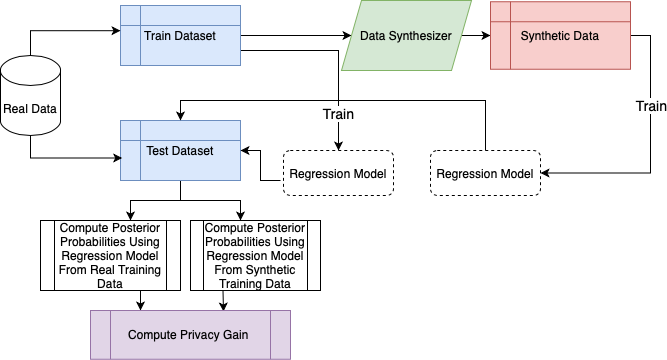}
    \caption{Attribute inference attack}
    \label{fig:MIA1}
  \end{minipage}
\end{figure}

To launch an attack, the attacker prepares two training datasets with and without the target record $t$ using the reference dataset $\mathcal{R}$, i.e. $\mathcal{R}$, and $\mathcal{R}\oplus t$ (see Fig.~\ref{fig:MIE}). Next, the attacker uses black-box access to the model for training two separate models on each dataset. The attacker then uses these to generate $s$ batches of synthetic data each consisting of $r$ rows, represented as $\mathcal{S}^{s}_{r}$. The synthetic batches are assigned a label of 0 and 1, respectively, based on the presence of $t$ in their training dataset. 
Thereafter, each batch of synthetic data is processed by a feature extraction method summarizing the information contained in each batch into a single vector. This is done in two ways: (i) naive extraction: computes the mean, median, and variance of every continuous column and the length of unique categories as well as the most and least frequently occurring category for every categorical column; (ii) correlation extraction: computes the pairwise correlations between all columns where the categorical columns are dummy-encoded.
This leads to the creation of a final dataset, containing an equal number of processed samples. This is split into train and test datasets. An attack model is trained on the training dataset and used to compute the privacy gain as $P_{Gain}=\frac{(P_{Real} - P_{Fake})}{2}$ where $P_{Fake}$ is the attack model's average probability of successfully predicting the correct label in the test-set and $P_{Real}=1$ since having access to the original training data ensures full knowledge of $t$'s presence~\cite{priv_mirage}. 

To conduct the membership inference evaluation, 4000 rows of real data are sampled from each dataset to form the reference dataset (i.e., $\mathcal{R}$) to train the synthetic models. Each batch for feature extraction is chosen to be of size $r=400$. And, $s=1200$ batches are generated such that the training dataset is of size 1000 with balanced number of classes. And, the test set contains 200 samples with balanced classes. To train the attack model we use the Random-Forest-Classifier. 
Results are averaged across 5 repetitions with 5 different targets $t$ for each dataset.
\\
\textbf{Attribute Inference Attack}~\cite{priv_mirage}. It is defined as a regression problem where the attacker attempts to predict the values of a sensitive target column provided he/she has black-box access to a generative model.

To launch an attribute inference attack and evaluate the privacy risk (see Fig.~\ref{fig:MIA1}), a dataset $\mathcal{R}$ sampled from the real distribution is split into training $\mathcal{R}_{Train}$ and testing $\mathcal{R}_{Test}$ datasets.  $\mathcal{R}_{Train}$ is fed into a generative model for generating a corresponding synthetic training dataset (i.e., $\mathcal{G}_{Train}$).
A linear regression model 
is then used to estimate the relationship between independent variables known to the attacker and the dependent sensitive variable for both $\mathcal{R}_{Train}$ and $\mathcal{G}_{Train}$. Then, to evaluate the privacy risk, the privacy gain is computed as: $P_{Gain} = \frac{(P_{Real}-P_{Fake})}{2}$, where $P_{Real}$ and $P_{Fake}$ denote the average posterior probabilities of correctly predicting the sensitive attribute on the real testing set given the linear models fitted on $\mathcal{R}_{Train}$ and $\mathcal{G}_{Train}$, respectively~\cite{priv_mirage}.  

To perform the attribute inference evaluation, we sample 5000 real rows from each dataset: 4900 samples are used as training dataset $\mathcal{R}_{Train}$) and 100 as testing dataset $\mathcal{R}_{Test}$. Moreover, we chose ``Age'', ``Age'' and ``Amount'' as sensitive attributes for the Adult, Loan and Credit datasets, respectively. Results are averaged across 5 repetitions.

\subsection{Results}
\label{results}

This section presents the results for all baselines based on the criteria established previously. Note that for measuring the statistical similarity and ML utility, the privacy budget $\epsilon$ is varied between 1 and 100 to study the influence of a strong vs weak privacy constraint, respectively. However, for evaluating the risk of privacy loss via membership and attribute inference attacks, a strict privacy budget of $\epsilon = 1$ is chosen as commonly used in prior work, e.g.~\cite{jordon2018pate}. This is done to thoroughly test experimentally the effectiveness of DP techniques offering strong theoretical guarantees. Note that $\delta=10^{-5}$ for all experiments and each table features the best results among the models with DP guarantees highlighted in bold. 
 
\subsubsection{ML Utility} 

\begin{table}[t]
\centering
\caption{Difference of accuracy (\%), F1-score, AUC and APR between original and synthetic data: average over 3 different datasets with different privacy budgets $\epsilon=1$ $\&$ $\epsilon=100$.}
\resizebox{0.95\columnwidth}{!}{
\begin{tabular}{|c||c|c|c|c||c|c|c|c|}
\hline
\multirow{2}{*}{\textbf{Method}} & \multicolumn{4}{c||}{\bf $\epsilon=1$} &
\multicolumn{4}{c|}{\bf $\epsilon=100$} \\
\cline{2-9}
 & \textbf{Accuracy} & \textbf{AUC}  & \textbf{APR}& \textbf{F1-Score} & \textbf{Accuracy} & \textbf{AUC}  & \textbf{APR} & \textbf{F1-Score} \\
\hline
{PATE-GAN}    & 10.8\% & \textbf{0.246} &   0.576 & 0.367 & 37.4\% & 0.416 &   0.566 & 0.412 \\
{DP-WGAN}     & \textbf{8.2\%}  & 0.408 &   0.58 & 0.368 & \textbf{10.8\%}  & 0.373 & 0.592 & 0.364 \\
{\systemd}    & 16.1\%&  0.302& \textbf{0.483} & \textbf{0.34} & 13\%    &  \textbf{0.265} & \textbf{0.475} &  \textbf{0.262}\\
{\systemg}    &  32.3\%& 0.377 &  0.604 & 0.454 &  13.7\% & 0.387  &  0.565 & 0.374 \\
\hline
{\tgan} & 2.6\% & 0.042 &  0.143 & 0.097 & 2.6\% & 0.042 &  0.143 & 0.097\\
\hline
\end{tabular}
}
\label{table:ML_allE}
\end{table}

Tab.~\ref{table:ML_allE} presents the results for the ML utility across the difference between models trained on original and synthetic data in the four chosen metrics. Ideally the difference should be zero.
\systemd achieves the best result in most cases beating both baselines and \systemg.
PATE-GAN surprisingly performs worse in terms of ML utility with a looser privacy budget than with a stricter privacy budget. This is mainly because the student discriminator is trained solely with generated samples of poor statistical similarity as found in Tab.~\ref{table:SS_allE}. Moreover, it is found 
that only the \systemd model consistently improves across all metrics with a looser privacy budget. It also showcases the best performance for both F1-score and APR metrics across all baselines and privacy budgets. This suggests that training the discriminator with DP guarantees, i.e. \systemd, is more optimal than training the discriminator with DP guarantees, i.e. \systemg. This is in line with the challenges faced by \systemg due to subsampling which hugely degrades performance by training multiple discriminators each using a smaller number of samples. Finally, the performance increase of \systemd in comparison to other baselines can be explained by its sophisticated neural network architecture (i.e., conditional GAN) and improved training objective (i.e., Wasserstein loss with gradient penalty). However, the cost of applying DP is noticeable by comparing to the DP-unaware \tgan.

\subsubsection{Statistical Similarity}

\begin{table}[t]
\centering 
\caption{Statistical similarity metrics averaged on 3 datasets with different privacy budgets $\epsilon=1$ $\&$ $\epsilon=100$.}
\resizebox{0.95\columnwidth}{!}{
\centering
\begin{tabular}{|c||c|c|c||c|c|c|}
\hline
\multirow{2}{*}{\textbf{Method}} & \multicolumn{3}{c||}{\bf $\epsilon=1$} &
\multicolumn{3}{c|}{\bf $\epsilon=100$} \\
\cline{2-7}
 & \textbf{Avg JSD} & \textbf{Avg WD} & \textbf{Diff. Corr.}  & 
\textbf{Avg JSD} & \textbf{Avg WD} & \textbf{Diff. Corr.}  \\
\hline
{PATE-GAN}    & 0.487   & 0.259   & 3.982  &  0.358  & 0.259   & 4.837\\
{DP-WGAN} &0.299   &  0.232 & 3.834 & 0.304 &  0.222   & 4.57\\
{\systemd}   & \textbf{0.246}  & \textbf{0.063}  &  4.168 & \textbf{0.127} &  \textbf{0.047} &  3.648\\
{\systemg}   &  0.376  &   0.189 &   \textbf{3.065}  &  0.389 & 0.174 &   \textbf{3.21} \\
\hline
{\tgan}   & 0.028  &  0.01 & 1.607 & 0.028  &  0.01 &   1.607 \\
\hline
\end{tabular}
}
\label{table:SS_allE}
\end{table}

The statistical similarity results are summarized in Tab.~\ref{table:SS_allE}.
Among all DP models \systemd is the only model which consistently improves across all three metrics when the privacy budget is increased. Similarly, \systemg sees an improvement across both the Avg-JSD and Avg-WD. The same is not true for PATE-GAN and DP-WGAN where DP-WGAN performs better across all metrics.
Moreover, they perform worse than the two variants of \system at both levels of epsilon. This highlights their inability to capture the statistical distributions during training despite a looser privacy budget. This is due to the lack of an effective training framework\rb{need justification or comment out}. Lastly, it is worth noting that \systemg features the best correlation distance at $\epsilon=1$ and $\epsilon=100$. It shows that reliably training the discriminator (with DP on the generator) is beneficial for capturing correlations in the data as compared to \systemd. Naturally, there is still a significant performance gap with respect to \tgan due to the application of DP.

\subsubsection{Resilience against Membership Inference Attack}
All DP baselines provide an empirical privacy gain close to $0.25$ for both feature extraction methods, see first two results column of Tab.~\ref{table:P_allE}. This indicates that differential private methods provide a strong privacy protection against membership attacks. It ensures that the average probability of success for any attack is close to the attacker's original prior, i.e $0.5$. \systemd and \systemg provide the highest resilience against a membership attack with naive and correlation feature extraction methods, respectively. Moreover, there is a clear decrease in the privacy gain achieved by \tgan showcasing that DP is needed to provide a stronger defense against membership inference attacks. 


\begin{table}[t]
\centering
\caption{Empirical privacy gain against membership attack with naive and correlation feature extraction, along with attribute inference attack: average over 3 different datasets with privacy budget $\epsilon=1$.} 
\resizebox{0.95\columnwidth}{!}{
\begin{tabular}{|c|c|c|c|}
\hline
\textbf{Method} & \textbf{Naive Privacy Gain} & \textbf{Correlation Privacy Gain} & \textbf{Attribute Inference Privacy Gain}\\
\hline
{PATE-GAN}    &  0.25   & 0.25 & \textbf{0.042} \\
{DP-WGAN}     &  0.255  & 0.256 & 0.04 \\
{\systemd}&  \textbf{0.266}  & 0.248 & 0.037 \\
{\systemg}&  0.245  & \textbf{0.26} & 0.038  \\
\hline
{\tgan}&  0.238  & 0.233 & 2e-4  \\
\hline
\end{tabular}
}
\label{table:P_allE}
\end{table}

\subsubsection{Resilience against Attribute Inference Attack}
The last column of Tab.~\ref{table:P_allE} shows the resilience against attribute inference attacks. PATE-GAN provides the greatest resilience, followed by DP-WGAN, \systemd, and \systemg. This is due to the superior quality of the synthetic data offered which enhances the attacker's probability of successfully inferring sensitive information.
Even if both variants of \system are less resilient than the two DP baselines, the difference with \tgan providing the worst/no resilience is still significant.
These results highlight the inherent trade-off between privacy and data utility i.e., increasing the utility directly worsens the privacy and vice versa. 

It is worth noting that the privacy gain for attribute inference attack for all baselines is close to 0. This suggests that the overall privacy protection offered against such attacks is quite low. However, it should be noted that the privacy gain is computed with respect to the real data. Thus, in case the real data itself provides a low probability of successfully inferring the correct target values for a sensitive attribute, then the synthetic dataset will perform in a similar manner resulting in a privacy gain close to 0.

\section{Conclusion}
\label{sec:conclusion}



Motivated by the risk of privacy leakage through synthetic tabular data, we propose a novel DP conditional wasserstein tabular GAN, \system. We rigorously analyze \system using it's two variants, namely \systemd and \systemg via the theoretical R\'enyi DP framework and elicit the privacy cost for additional losses used by the generator to enhance data quality. Moreover, we empirically showcase the data utility achieved by applying DP-SGD to train the discriminator vs generator, respectively, Additionally, we rigorously evaluate the privacy robustness against practical membership and attribute inference attacks. 

Our results on three tabular datasets show that synthetic tabular data generated by DP-SGD achieves higher data utility as compared to the PATE framework. Moreover, we find that \systemd outperforms \systemg, illustrating that the discriminator trained with DP guarantees is more optimal under stringent privacy budgets. Finally, in terms of data utility and reliance to privacy attacks , \systemd improves upon prior work by 18\% across 4 ML models in terms of the average precision score and all DP baselines reduce the success rate of membership attacks by approx. 50\%. Therefore, this work showcases the effectiveness of DP for protecting the privacy of sensitive datasets being used for training tabular GANs. However, further enhancement of the quality of synthetic data at strict privacy budgets (i.e., $\epsilon \leq 1$) \rb{why $leq$ ?} is still needed. Ultimately, there is an inherent trade-off between privacy and utility and obtaining the most optimal balance between both is left for future work.


\bibliography{acml21_DP}

\appendix





\end{document}